\documentclass[review,11pt]{ReportTemplate}
\usepackage{bm}
\usepackage{graphicx}
\usepackage{paralist, algorithmic, algorithm}
\usepackage{amsmath, mathrsfs, amssymb}
\usepackage{subfigure}
\usepackage{geometry}
\usepackage{amsmath, amssymb}
\usepackage{multirow}
\usepackage{arydshln}  
\usepackage{longtable, booktabs}
\usepackage{float}
\usepackage[permil]{overpic}
\usepackage{color, xcolor}
\usepackage{helvet}   
\usepackage{setspace}
\usepackage{natbib}   
\usepackage{caption}  
\usepackage{newfloat}
\usepackage{listings}

\newcommand\bs{\boldsymbol}
\newcommand\bb{\mathbb}
\newcommand\al{\mathcal}

\definecolor{steelblue}{RGB}{46, 82, 180}

\begin{document}
\begin{frontmatter}
\title{Multi-Instance Partial-Label Learning: Towards Exploiting Dual Inexact Supervision}

\author{Wei Tang, Weijia Zhang, and Min-Ling Zhang\corref{cor1}}
\address{School of Computer Science and Engineering, Southeast University, Nanjing 210096, China\\
Key Laboratory of Computer Network and Information Integration \\(Southeast University), Ministry of Education, China \\
\rm{\{tangw, zhangwj, zhangml\}@seu.edu.cn}
}
\cortext[cor1]{ Corresponding author.}

\begin{abstract} 
Weakly supervised machine learning algorithms are able to learn from ambiguous samples or labels, e.g., multi-instance learning or partial-label learning. However, in some real-world tasks, each training sample is associated with not only multiple instances but also a candidate label set that contains one ground-truth label and some false positive labels. Specifically, at least one instance pertains to the ground-truth label while no instance belongs to the false positive labels. In this paper, we formalize such problems as \emph{multi-instance partial-label learning} (MIPL). Existing multi-instance learning algorithms and partial-label learning algorithms are suboptimal for solving MIPL problems since the former fail to disambiguate a candidate label set, and the latter cannot handle a multi-instance bag. To address these issues, a tailored algorithm named {M\scriptsize{IPL}}G{\scriptsize{P}}, i.e., \emph{Multi-Instance Partial-Label learning with Gaussian Processes}, is proposed. {M\scriptsize{IPL}}G{\scriptsize{P}} first assigns each instance with a candidate label set in an augmented label space, then transforms the candidate label set into a logarithmic space to yield the disambiguated and continuous labels via an exclusive disambiguation strategy, and last induces a model based on the Gaussian processes. Experimental results on various datasets validate that {M\scriptsize{IPL}}G{\scriptsize{P}} is superior to well-established multi-instance learning and partial-label learning algorithms for solving MIPL problems. Our code and datasets will be made publicly available.
\end{abstract} 

\end{frontmatter}

\section{Introduction}
\label{sec:introduction}
In standard supervised learning, each training sample is represented by a single instance associated with a class label. In recent years, supervised learning has achieved fruitful progress when a large amount of supervision is available. However, annotating large amounts of high-quality labels is time-consuming and costly, especially in fields that require expert knowledge. To overcome these issues, several weakly supervised learning paradigms are proposed and have attracted significant research attention.   
\begin{figure}[!b]
\centering
\begin{overpic}[width=120mm]{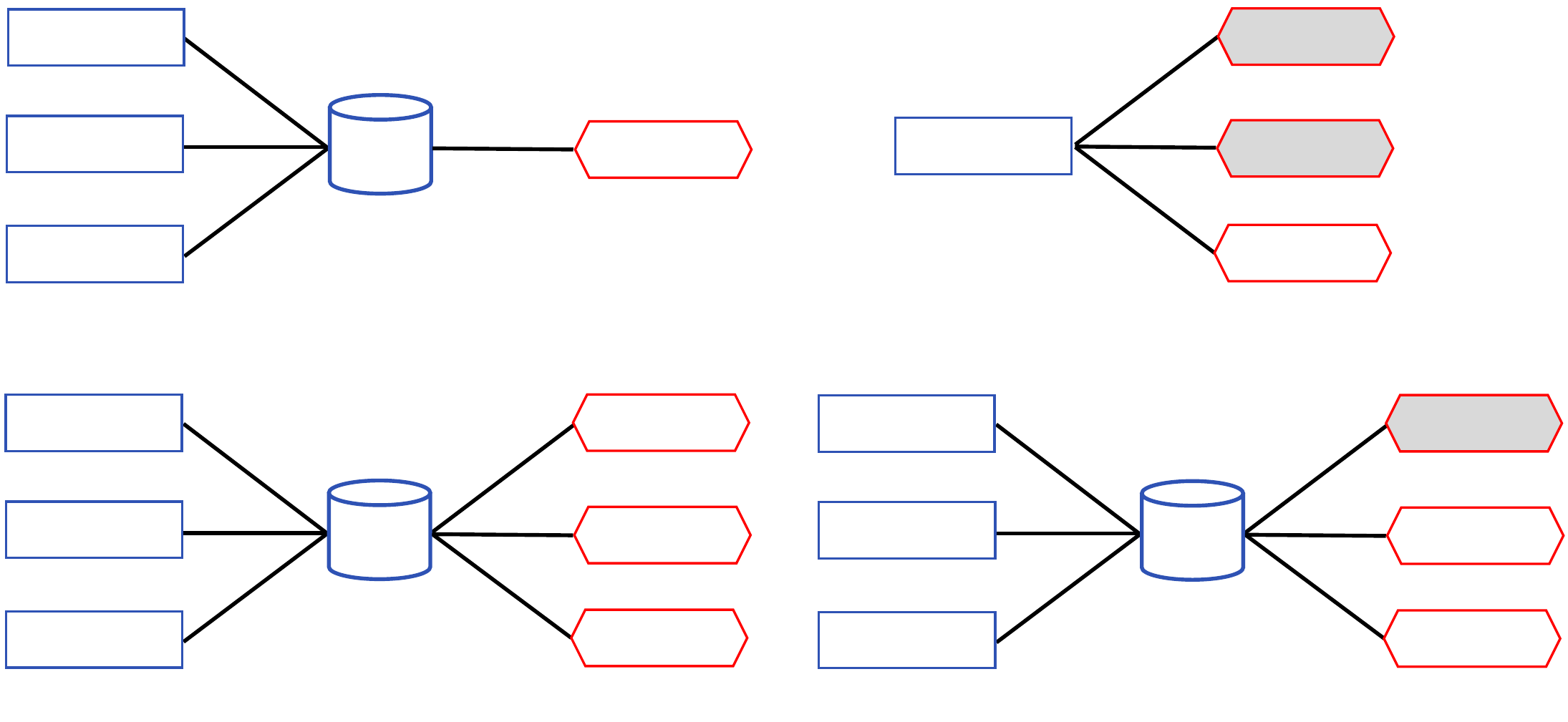}
\put(10, 430) {\small  \textcolor{steelblue}{instance}}   
\put(10, 360) {\small  \textcolor{steelblue}{instance}}   
\put(10, 290) {\small  \textcolor{steelblue}{instance}}  
\put(220, 355) {\small  \textcolor{steelblue}{bag}}   
\put(392, 357) {\small \textcolor{red}{label}}
\put(70, 250) {\small (a) Multi-instance learning}

\put(576, 360) {\small \textcolor{steelblue}{instance}}   
\put(803, 428) {\small \textcolor{red}{label}}
\put(803, 357) {\small \textcolor{red}{label}}
\put(803, 290) {\small \textcolor{red}{label}}
\put(570, 250) {\small (b) Partial-label learning}

\put(9, 182) {\small  \textcolor{steelblue}{instance}}   
\put(9, 113) {\small  \textcolor{steelblue}{instance}}   
\put(9, 43) {\small  \textcolor{steelblue}{instance}}  
\put(220, 110) {\small  \textcolor{steelblue}{bag}}   
\put(392, 182) {\small \textcolor{red}{label}}
\put(392, 110) {\small \textcolor{red}{label}}
\put(392, 44) {\small \textcolor{red}{label}}
\put(-5, 0) {\small (c) Multi-instance multi-label learning}

\put(527, 182) {\small  \textcolor{steelblue}{instance}}   
\put(527, 113) {\small  \textcolor{steelblue}{instance}}   
\put(527, 43) {\small  \textcolor{steelblue}{instance}}  
\put(740, 110) {\small  \textcolor{steelblue}{bag}}   
\put(910, 182) {\small \textcolor{red}{label}}
\put(910, 110) {\small \textcolor{red}{label}}
\put(910, 44) {\small \textcolor{red}{label}}
\put(500, 0) {\small (d) Multi-instance partial-label learning}
\end{overpic}
\caption{Different weakly supervised learning frameworks, where the grey polygons refer to the false positive labels.}
  \label{fig:fourML}
\end{figure}
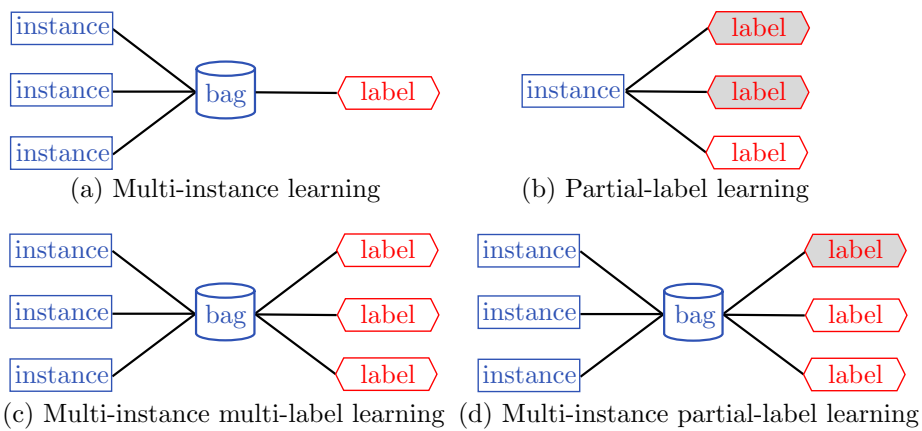

According to the quality and number of the labels, weak supervision can be roughly divided into three categories, i.e., incomplete, inexact, and inaccurate supervision \cite{zhou2018brief}. The inexact supervision refers to coarse-grained labels and contains two popular learning frameworks, i.e., multi-instance learning (MIL) and partial-label learning (PLL). 
As illustrated in Figure \ref{fig:fourML}a, in MIL, multiple training instances are arranged in a bag and we only know the binary bag-level label rather than the instance-level labels \cite{amores2013multiple, carbonneau2018multiple}. Although the bag-level label is known, the exact labels for the instances in the bag are ambiguous.
The framework of PLL is shown in Figure \ref{fig:fourML}b, where each training sample is represented by a single instance coupled with a candidate label set, which consists of a ground-truth label and several false positive labels \cite{JinG02}. Therefore, the mapping from the instance to the concealed ground-truth label is ambiguous. In a sense, multi-instance learning and partial-label learning are dual frameworks to each other in which inexact supervision exists in the instance space and the label space, respectively.

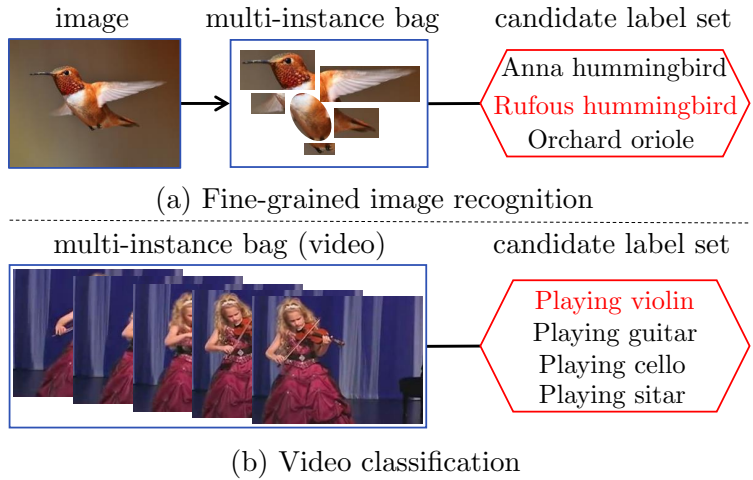
\begin{figure}[!t]
\centering
\begin{overpic}[width=100mm]{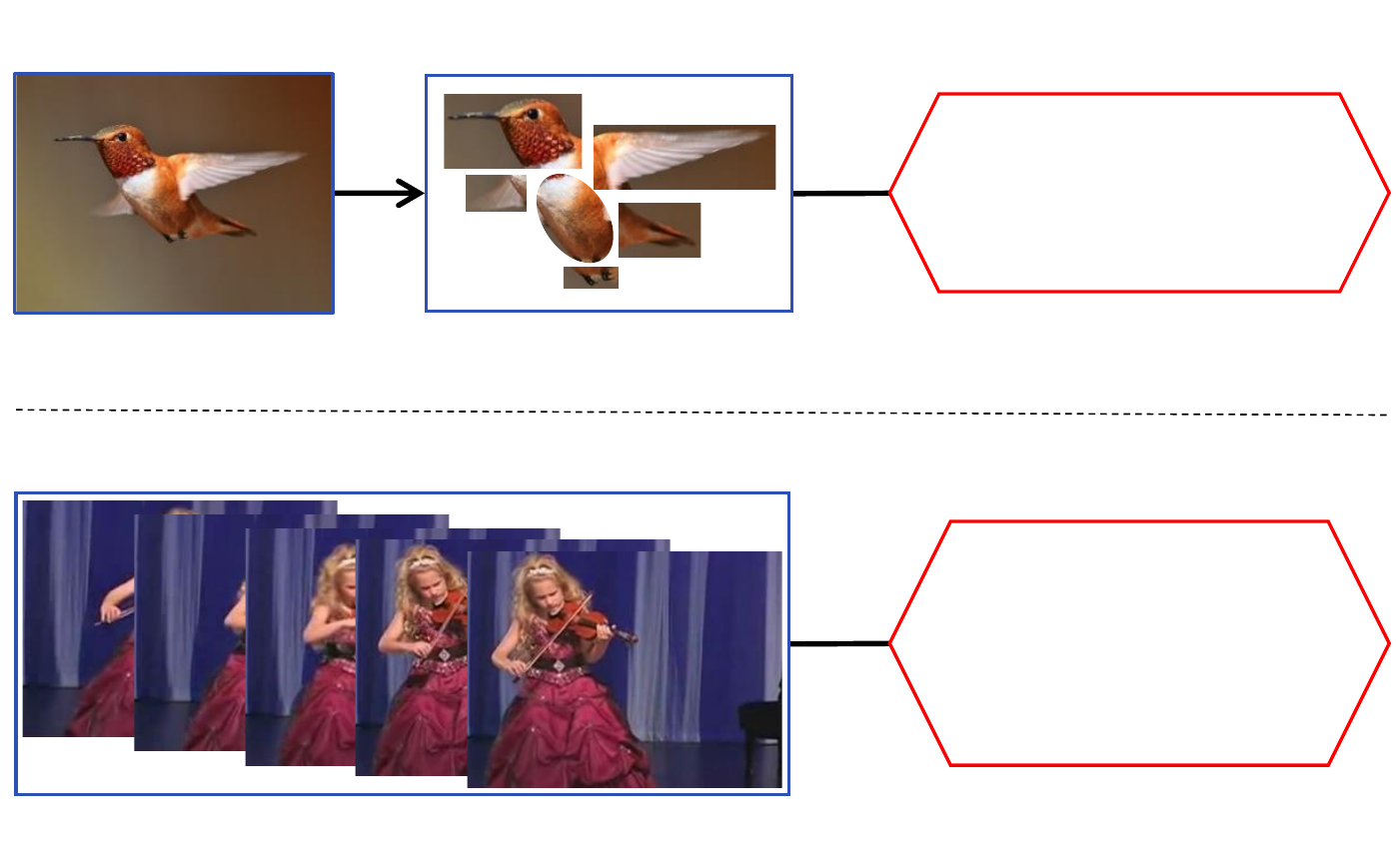}
\put(72, 585) {image}
\put(272, 585) {multi-instance bag}
\put(657, 585) {candidate label set}
\put(665, 520) {\small Anna hummingbird}  
\put(655, 470) {\small \textcolor{red}{Rufous hummingbird}}
\put(700, 425) {\small Orchard oriole}  
\put(205, 345) {(a) Fine-grained image recognition}

\put(70, 285) {multi-instance bag (video)}
\put(655, 285) {candidate label set}
\put(710, 210) {\small \textcolor{red}{Playing violin}}
\put(707, 168) {\small Playing guitar}  
\put(715, 125) {\small Playing cello}  
\put(715, 85) {\small Playing sitar}  
\put(305, -5) {(b) Video classification}
\end{overpic}
\caption{Potential applications of MIPL, where the red is the ground-truth label.}
  \label{fig:example}
\end{figure}

However, ambiguities can exist simultaneously in the instance space and the label space. For example, in fine-grained image recognition (as illustrated in Figure \ref{fig:example}a), each image can be treated as a multi-instance bag \cite{XuHZT15}.
As the supervision is provided by noisy web labels, the label set of each image contains not only a ground-truth label but also false positive labels \cite{xu22webly}. Therefore, we can assign each multi-instance bag with a candidate label set rather than an exact label, and train a model to learn from the partially labeled multi-instance bags. 
In video classification (as illustrated in Figure \ref{fig:example}b), each video consists multiple frames represented as a set of instances, and the labels from social media contain noises that need to be corrected manually \cite{GhadiyaramTM19}. 
The labeling cost can be significantly reduced if the video classification algorithm can learn from samples represented as sets of instances associated with candidate label sets.

Motivated by the potential applications, we formalize a novel framework named multi-instance partial-label learning (MIPL), which handles ambiguities in the instance space and the label space simultaneously. In Figure \ref{fig:fourML}d, each training sample is represented by a multi-instance bag associated with a bag-level candidate label set, which consists of one ground-truth label and some false positive labels. Moreover, the bag contains at least one instance that belongs to the ground-truth label while no instance pertains to the false positive labels. Therefore, inexact supervision exists both in the instance space and the label space in MIPL. It is noteworthy that multi-instance partial-label learning is different from multi-instance multi-label learning (MIML) presented in Figure \ref{fig:fourML}c, where each multi-instance bag is also associated with a label set \cite{ZhouZ06}. The differences between MIPL and MIML lie in that the label set in MIML only contains ground-truth labels, while the candidate label set in MIPL consists of one ground-truth label and some false positive labels.

To solve the MIPL problems, we propose a tailored algorithm named {M\scriptsize{IPL}}G{\scriptsize{P}}, i.e., \emph{Multi-Instance Partial-Label learning with Gaussian Processes}. First, in order to assign each instance with a candidate label set containing the ground-truth label, we propose a label augmentation strategy to augment each candidate label set with a negative class label. Second, by virtue of the Dirichlet disambiguation strategy, {M\scriptsize{IPL}}G{\scriptsize{P}} transforms the augmented candidate labels to the disambiguated and continuous labels and construct a Gaussian likelihood in a logarithmic space. Last, to infer the parameters of the Dirichlet disambiguation strategy accurately, {M\scriptsize{IPL}}G{\scriptsize{P}} induces a multi-output Gaussian processes regression model with GPU accelerations.

Empirical evaluation of {M\scriptsize{IPL}}G{\scriptsize{P}} is conducted on five MIPL datasets. The experimental results indicate that:   
(a) The MIPL is an exclusive problem that is difficult to be solved by neither multi-instance learning approaches nor partial-label learning approaches.
(b) {M\scriptsize{IPL}}G{\scriptsize{P}} achieves superior results against well-established multi-instance learning and partial-label learning approaches.
(c) The proposed label augmentation and Dirichlet disambiguation strategies are both important for solving the MIPL problem.
 
The rest of the paper is organized as follows. First, related work is briefly reviewed. Second, we present the proposed {M\scriptsize{IPL}}G{\scriptsize{P}} and report the experimental setting and results. Last, we conclude this paper.

\section{Related Work}
\label{sec:related_work}
\subsection{Multi-Instance Learning}
Multi-instance learning algorithms can be roughly divided into two groups, i.e., instance-level algorithms and bag-level algorithms \cite{amores2013multiple}. The former predict a bag-level label by aggregating instance-level ones, e.g., maximizing or averaging the probabilities of all instances in a bag. The latter induce a classifier by treating each bag as a whole entity, which includes the bag-space paradigm and the embedded-space paradigm.

In general, probabilistic multi-instance methods create a model that characterizes the distribution of instance-level labels and yields aggregated bag-level labels.
\citet{kim10_gpmil} proposes a nonparametric model to capture the underlying generative process by integrating a special bag class likelihood into the Gaussian processes. Along this line, \citet{HauBmannHK17} modifies the standard bag likelihood and infers an instance-label Gaussian processes classifier using variational Bayes. To model the dependencies among the instances, the variational autoencoder is employed to predict both the instance-level and bag-level labels \cite{Weijia21, Weijia22}. 
A recent tendency to address MIL problems is combining neural networks with the attention mechanism, where the attention scores indicate the importance of the instances to the bag \cite{IlseTW18, wu2021combining, zhang2022dtfd}. To our knowledge, these MIL methods are designed for binary classification problems, which cannot be directly adopted to solve MIPL problems. 
Although there are some multi-instance learning algorithms that can handle multi-classification problems \cite{ShaoBCWZJZ21, BrandBES021}, they cannot tackle the challenge of false positive labels in the candidate label set.

\subsection{Partial-Label Learning}
Partial-label learning algorithms utilize identification-based or average-based disambiguation strategies to disambiguate the candidate label sets. The average-based disambiguation strategy treats all labels in the candidate label set equally, and averages the output of the model to achieve disambiguation \cite{cour2011learning, GongLTYYT18}. The identification-based disambiguation strategy considers the potential ground-truth label as a latent variable, and disambiguates the ambiguous labels by optimizing the objective function related to the latent variable \cite{yu2016maximum, feng2019partial}. 

Based on the graphic model, \citet{JinG02} minimizes relative entropy between the estimated label distribution and the prior distribution of the class labels. To capture underlying structures of the data, \citet{liu2012conditional} maps training instances to mixture components and samples a label for each mixture component. Based on the Gaussian processes, \citet{zhou2016partial} defines a non-Gaussian likelihood to disambiguate the candidate label sets and computes the posterior distribution using Laplace approximation. Recently, some deep learning-based disambiguation methods have been investigated for partial-label learning \cite{LvXF0GS20, wang2022contrastive}. However, we note that all of them cannot handle multi-instance bags.

\section{Methodology}
\label{sec:methodology}
In this section, we propose a MIPL algorithm based on Gaussian processes, i.e., {M\scriptsize{IPL}}G{\scriptsize{P}}. To the best of our knowledge, this is the first algorithm to address the MIPL problems.
First, we introduce the notations and define an augmented label space, which equips each instance among a bag with a suitable candidate label set. Then, we propose a novel Dirichlet disambiguation strategy that effectively disambiguates the candidate label sets. Last, we present a multi-output Gaussian process model.	

Let $\al{X} = \mathbb{R}^d$ denote the instance space and $\al{Y} = \{l_1, l_2,\cdots, l_q\}$ denote the label space with $q$ class labels. The goal of MIPL is to learn a classifier $h:2^{\al{X}} \to \al{Y}$ from a training dataset $\{(\bs{X}_i,\bs{y}_i)\mid 1 \le i \le m \}$\footnote{Unless otherwise stated, we use symbols in bold to denote matrices and vectors, and use regular symbols to denote scalars.} with $m$ bags and corresponding candidate label sets. Specifically, a multi-instance partial-label sample is denoted as $(\bs{X}_i, \bs{y}_i)$, where $\bs{X}_i = [\bs{x}_i^1, \bs{x}_i^2, \cdots, \bs{x}_i^{z_i}]^\top$ is a bag of $z_i$ instances, $\bs{x}_i^j \in \al{X}$ for $\forall j \in \{1,2,\cdots,z_i\}$, and $\bs{y}_i = [y_i^1,y_i^2,\cdots,y_i^q]^\top \in \{0,1\}^q$ is the candidate label set of $\bs{X}_i$ where $y_i^c = 1$ means that the $c$-th label is one of the candidate labels of $\bs{X}_i$ and $y_i^c = 0$ otherwise. 

\subsection{Label Augmentation}
In multi-instance learning, only the bag-level labels are available, while the instance-level labels are unknown. To tackle this problem, a straightforward approach is to propagate the bag label to be the dummy label of all instances in the bag. An obvious problem with this approach is that it will incorrectly assign the negative instances in a positive bag with positive labels. 
Analogously, in MIPL, if the bag-level candidate label set is directly applied to all the instances in the bag, the ground-truth labels of a substantial amount of instances will not exist in their candidate label sets, which violates the settings of partial-label learning.

To address the issue, we propose to utilize an augmented label space $\widetilde{\al{Y}} = \{l_1, l_2,\cdots, l_q, l_{neg}\}$ with $\tilde{q}=q + 1$ class labels, which augments a negative class label to the original label space $\al{Y}$. Specifically, we assign instances that do not pertain to label space $\al{Y}$ with the augmented negative class $l_{neg}$.
For example, given a multi-instance bag $\bs{X}_i$ associated with a candidate label set $\bs{y}_i = [y_i^1,y_i^2, \cdots, y_i^q]^\top$, each instance in $\bs{X}_i$ is endowed with an augmented candidate label set $\widetilde{\bs{y}}_i= [y_i^1,y_i^2,\cdots,y_i^q, y_i^{neg}]^\top$ where $y_i^{neg} = 1$.

Consequently, we can derive the instance-level features and semantic information based on the augmented label space. Let ${\bf{X}} = [\bs{x}_1^1, \bs{x}_1^2, \cdots, \bs{x}_1^{z_1}, \bs{x}_2^1, \bs{x}_2^2, \cdots, \bs{x}_m^{1}, \cdots,\bs{x}_m^{z_m}]^\top \in \bb{R}^{n \times d}$ denote the feature matrix of $n$ instances spread over $m$ bags and $\widetilde{\bf{Y}} = [\widetilde{\bs{y}}_1^1, \widetilde{\bs{y}}_1^2,\cdots, \widetilde{\bs{y}}_1^{z_1}, \widetilde{\bs{y}}_2^1, \\ \widetilde{\bs{y}}_2^2, \cdots,\widetilde{\bs{y}}_m^1, \cdots, \widetilde{\bs{y}}_m^{z_m}]^\top \in \bb{R}^{n \times \tilde{q}}$ denote the partial-label matrix of the $n$ instances, where $n = \sum_{i=1}^{m}z_i$ is the total number of the instances in the dataset and $\widetilde{\bs{y}}_i = \widetilde{\bs{y}}_i^1 = \widetilde{\bs{y}}_i^2 = \cdots = \widetilde{\bs{y}}_i^{z_i}$ is the augmented candidate label set of $\bs{X}_i$ as well as all instances in bag $\bs{X}_i$. Notably, the label augmentation occurs in the data processing phase, which does not increase training overhead.

\subsection{Dirichlet Disambiguation}
Motivated by the disambiguation strategies in partial-label learning, we conceive a novel disambiguation strategy for MIPL, which is named \emph{Dirichlet disambiguation}.

Given an augmented MIPL training dataset $(\bf{X, \widetilde{\bf{Y}}})$ with $m$ bags totalling $n$ instances, one instance and its candidate label set can be written as $( \bs{x}_i^j, \widetilde{\bs{y}}_i^j)$ for $\forall i \in \{1,2,\cdots,m\} \text{~and~} j \in \{1,2, \cdots, z_i\}$. When the context is clear, we omit the instance index $j$ to $(\bs{x}_i, \widetilde{\bs{y}}_i)$ for brevity.
It is intuitive to use a categorical distribution $\text{Cat}(\bs{\theta}_i)$ to infer the ground-truth label of the instance, where the class probability $\bs{\theta}_i = [\theta_i^1, \theta_i^2, \cdots, \theta_i^q, \theta_i^{neg}]^\top$ is a multivariate continuous random variable constrained in a $q$ dimensional probability simplex, i.e., $\sum_{c=1}^{\tilde{q}} \theta_i^c = 1$ and $\theta_i^c \geq 0$ ($\forall c \in \{1,2, \cdots, q, neg\}$). It is worth noting that the simplex promotes mutual exclusion among the candidate labels.
In order to establish the class probability of the categorical distribution, we utilize the Dirichlet distribution, which is the conjugate prior to the categorical distribution, to reduce computational difficulty. 
Accordingly, the Dirichlet distribution $\text{Dir}({\bs{\alpha}}_i)$ with a concentration parameter $\bs{\alpha}_i = [\alpha_i^1, \alpha_i^2, \cdots, \alpha_i^q, \alpha_i^{neg}]^\top$ is adopted to measure $\bs{\theta}_i$. Concretely, the likelihood model is given by:
\begin{equation}
\label{eq:likelihood}
	p(\bs{\widetilde{y}}_i \mid \bs{\alpha}_i) = \text{Cat}(\bs{\theta}_i) \text{,~~} \bs{\theta}_i \sim \text{Dir}({\bs{\alpha}_i}).
\end{equation}

To draw the coefficient $\bs{\theta}_i$ from the Dirichlet distribution, the accurate value of $\bs{\alpha}_i$ becomes pivotal. In supervised learning, each instance is associated with a unique ground-truth, and thus a constant weight $w$ can be directly added to the index corresponding to the ground-truth. For example, given an observation $\bs{y}_i = [y_i^1,y_i^2, \cdots, y_i^q]^\top$ that satisfies $y_i^c=1$ and $y_i^j = 0$ ($\forall j \ne c$), we have $\alpha_i^c = w + \alpha_{\epsilon}$ and $\alpha_i^j = \alpha_{\epsilon}$ ($\forall j \ne c$), where $\alpha_{\epsilon}$ is the Dirichlet prior such that $0 < \alpha_{\epsilon} \ll 1$.
However, it is not appropriate to add a constant weight directly in MIPL, since the candidate label set is contaminated by the false positive labels. To overcome this limitation, we propose to synergize the Dirichlet distribution with an iterative disambiguation strategy to identify the ground-truth label from the contaminated candidate label set. To achieve the disambiguation strategy, $\bs{\alpha}_i = [\alpha_i^1, \alpha_i^2, \cdots, \alpha_i^{q}, \alpha_i^{neg}]$ is initialized with uniform weights for $c \in \{1,2, \cdots, q, neg\}$:
\begin{equation}
\label{eq:init_weight}
\alpha_{i}^c=\left\{\begin{array}{cc}
	\frac{1}{\left|\widetilde{\bs{y}}_i\right|} + \alpha_{\epsilon} & \text {if } y_i^c = 1, \\
	\alpha_{\epsilon} & \text {otherwise, }
\end{array}\right.
\end{equation}
where $0 < \alpha_{\epsilon} \ll 1$ and $|\widetilde{\bs{y}}_i|$ is the cardinality which measures the number of non-zero elements in the augmented candidate label set $\widetilde{\bs{y}}_i$. The softmax value of the classifier output $\widetilde{\bs{h}}_i = \widetilde{\bs{h}}(\bs{x}_i)=[h_i^1, h_i^2, \cdots, h_i^q, h_i^{neg}]^\top$ on the candidate label set indicates the probability that each candidate label is a ground-truth label. Therefore, we utilize the softmax value to gradually eliminate the false positive labels and identify the ground-truth label in each iteration:
\begin{equation}
\label{eq:update_weight}
\alpha_{i}^c=\left\{\begin{array}{cc}
	\frac{\exp (h_i^c)}{\sum_{y_i^t =1} \exp (h_i^t)} + \alpha_{\epsilon} & \text {if } y_i^c = 1, \\
	\alpha_{\epsilon} & \text {otherwise.}
\end{array}\right.
\end{equation}

Next, the problem becomes how to sample from the Dirichlet distribution. Considering both generation quality and cost, we design a two-step process to generate the Dirichlet samples from $\tilde{q}$ independent Gamma-distributed random variables. First, we generate $\tilde{q}$ Gamma-distributed random variables $\{\gamma_i^1, \gamma_i^2, \cdots, \gamma_i^q, \gamma_i^{neg}\}$ from the Gamma distribution $\text{Gamma}(\alpha_i^c, 1)$ for $c = 1,2, \cdots, q, neg$, respectively. Then, we normalize the $\tilde{q}$ Gamma-distributed random variables to obtain the realizations. The formulations of the generation process are as follows:
\begin{equation}
\label{eq:gamma}
	\theta_i^c = \frac{\gamma_i^c}{\sum_{j=1}^{\tilde{q}} \gamma_i^j} \text{,~~} \gamma_i^c \sim \text{Gamma}(\alpha_i^c, 1). 
\end{equation}
The probability density function of $\text{Gamma}(\alpha_i^c, 1)$ is $\frac{\gamma^{(\alpha_i^c - 1)} \exp (-\gamma)}{\Gamma(\alpha_i^c)}$, where $\alpha_i^c > 0$ is called the shape parameter and $\Gamma(\cdot)$ is the gamma function.

In order to use an exact Gaussian processes model to infer $\alpha_i^c$ accurately,
we employ the random variables $\dot{x}_i^c$ drawn from a logarithmic normal distribution $\text{LogNormal} (\dot{y}_i^c, \dot{\sigma}_i^c)$ to approximate the Gamma-distributed random variables $\gamma_i^c$ by moment matching, i.e., mean matching $\bb{E}[\gamma_i^c] = \bb{E}[\dot{x}_i^c]$ and variance matching $\bb{V}[\gamma_i^c] = \bb{V}[\dot{x}_i^c]$:
\begin{equation}
\label{eq:moment}
\begin{array}{cc}
	\alpha_i^c =  \exp  \left(\dot{y}_i^c +\frac{\dot{\sigma}_i^c}{2} \right), \\
	\alpha_i^c =\left(\exp \left(\dot{\sigma}_i^c\right)-1\right) \exp \left(2 \dot{y}_i^c + \dot{\sigma}_i^c\right).
\end{array}
\end{equation}
As Milios et al. \cite{MiliosCMRF18} shows, it is reasonable to estimate $\gamma_i^c$ by $\text{LogNormal} (\dot{y}_i^c, \dot{\sigma}_i^c)$. The parameters of $\text{LogNormal} (\dot{y}_i^c, \dot{\sigma}_i^c)$ are derived by solving Eq. (\ref{eq:moment}):
\begin{equation}
\label{eq:new_label}
\begin{array}{cc}
	\dot{\sigma}_i^c = \log \left(\frac{1}{\alpha_i^c} + 1 \right), \\
	\dot{y}_i^c =  \log \alpha_i^c - \frac{\dot{\sigma}_i^c}{2} = \frac{3}{2} \log \alpha_i^c - \frac{1}{2} \log (\alpha_i^c + 1),
\end{array}
\end{equation}
where $\dot{y}_i^c$ is a continuous label in the logarithmic space and $\dot{\sigma}_i^c$ is a variance related to $\dot{y}_i^c$.
Based on the above Dirichlet disambiguation strategy, the original sample $(\bs{x}_i, \widetilde{\bs{y}}_i)$ is transformed into $(\bs{x}_i, \dot{\bs{y}}_i)$, where $\dot{\bs{y}}_i$ is a candidate label set with continuous labels. Meanwhile, a Gaussian likelihood is constructed in the logarithmic space.
Given a MIPL training dataset $({\bf{X}}, \widetilde{{\bf{Y}}})$, we reshape the transformed candidate label matrix to yield a row-wise concatenation $\dot{{\bf{Y}}} = [\dot{\bs{y}}_1^1; \dot{\bs{y}}_1^2; \cdots; \dot{\bs{y}}_1^{z_1}; \dot{\bs{y}}_2^1; \dot{\bs{y}}_2^2; \cdots; \dot{\bs{y}}_m^1, \cdots; \dot{\bs{y}}_m^{z_m}] \in \bb{R}^{\tilde{q}n}$.

\subsection{Gaussian Processes Regression Model}
Based on the continuous candidate label set matrix $\dot{{\bf{Y}}}$, we can transform the MIPL from a multi-class classification problem to a Gaussian processes regression problem with $\tilde{q}$ outputs. To estimate $\bs{\alpha}_i$ accurately, we develop a multi-output Gaussian processes regression model for MIPL based on the Gaussian likelihood in the logarithmic space.

For the multi-output Gaussian processes regression model, the vector of $\tilde{q}$ latent functions $\{f^1(\cdot), f^2(\cdot), \cdots, f^{q}(\cdot), f^{neg}(\cdot)\}$ at all $n$ training instances is first introduced: ${\bf{F}} = [\bs{f}^1, \bs{f}^2, \cdots, \bs{f}^{neg}]^\top = [f_1^1, \cdots, f_n^1, f_1^2, \cdots, f_n^2, \cdots, f_1^{neg}, \cdots, f_n^{neg}]^\top$, where the latent variate ${\bf{F}}$ has length $\tilde{q}n$. 
The distribution of ${\bf{F}}$ is defined by a prior mean function ${\bf{\mu}}= {\bf{0}}$ and a covariance function, i.e., a prior kernel $k(\cdot,\cdot): \bb{R}^d \times \bb{R}^d \to \bb{R}$, which is chosen to be a Mat{\'{e}}rn kernel \cite{RasmussenW06} in this paper.  
For $\forall c, c^\prime \in \{1, 2, \cdots,q,neg\}$, the correlation of the outputs at any a pair of instances $\bs{x}$ and $\bs{x}^\prime$ can be represented as:
\begin{equation}
\label{eq:kernel}
\begin{array}{l}
\operatorname{Cov}[f^c(\bs{x}), f^{c^\prime}(\bs{x^\prime})] = k^{c}\left(\bs{x}, \bs{x}^{\prime}\right) = \left\{\begin{array}{cc}
\frac{2^{1 - \nu}}{\Gamma(\nu)}   \left( \frac{\sqrt{2 \nu} d}{\ell} \right)^{\nu} K_\nu \left( \frac{\sqrt{2 \nu} d}{\ell} \right) & \text {if~} c=c^\prime \text{,}\\
0 & \text{otherwise.}
\end{array}\right.
\end{array}
\end{equation}
Generally speaking, $\nu$ is a smoothness parameter and takes the value from $\{0.5, 1.5, 2.5\}$. Here, $\ell$ is a positive parameter, $d$ is the Euclidean distance between $\bs{x}$ and $\bs{x}^\prime$, and $K_\nu$ is a modified Bessel function.
Finally, the covariance matrix ${\bf{K}} \in \bb{R}^{\tilde{q}n \times \tilde{q}n}$ of the $\tilde{q}$ latent processes is block diagonal by the matrices ${\bf{K}}^1, {\bf{K}}^2, \cdots, {\bf{K}}^{\tilde{q}}$ of shape $n \times n$.
Gaussian processes place a Gaussian prior over latent variable ${\bf{F}} \sim \al{GP}(\bf{0, K})$, i.e., $P({\bf{F}} \mid {\bf{X}}) = \al{N}(\bf{0, K})$, and the Gaussian likelihood in the logarithmic space is $P(\dot{\bf{Y}} \mid {\bf{F}}) = \al{N}({\bf{F}}, \Sigma)$, where $\Sigma$ is the matrix form of $\dot{\sigma}_i^c$ in $\text{LogNormal} (\dot{y}_i^c, \dot{\sigma}_i^c)$. Following the Bayes' rule, the posterior distribution $P({\bf{F}} \mid {\bf{X}}, \dot{\bf{Y}}) \propto P({\bf{F}} \mid {\bf{X}}) P(\dot{\bf{Y}} \mid {\bf{F}})$ and the marginal likelihood $P(\dot{\bf{Y}} \mid {\bf{X}}) = \int_{\bs{F}} P({\bf{F}} \mid {\bf{X}}) P(\dot{\bf{Y}} \mid {\bf{F}})$ are both Gaussian. 
\begin{figure}[!t]
\centering
\begin{overpic}[width=100mm]{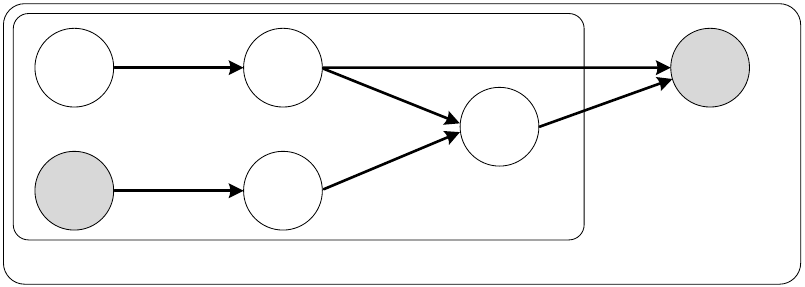}
\put(75, 265) {$\alpha_i^j$}
\put(75, 110) {$x_i^j$}
\put(335, 260) {$\theta_i^j$}
\put(335, 110) {$f_i^j$}
\put(605, 190) {$y_i^j$}
\put(865, 265) {$\bs{y}_i$}
\put(430, 75) {$j \in \{1, 2, \cdots, z_i\}$}
\put(700, 20) {$i \in \{1, 2, \cdots, m\}$}
\end{overpic}
 \caption{Plate diagram for {M\scriptsize{IPL}}G{\scriptsize{P}}, where the grey circles and white ones are the observed variables and latent variables, respectively.}
  \label{fig:diagram}
\end{figure}
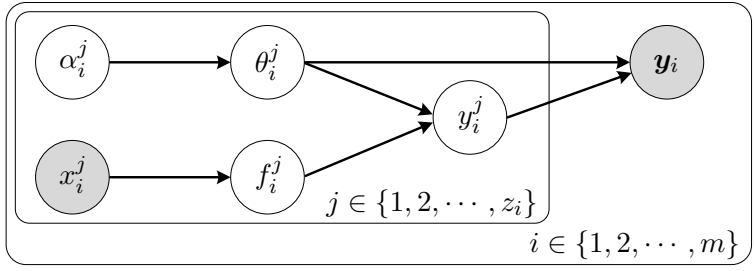

The above likelihoods are based on the instances-level features and labels. In multi-instance learning, however, a ubiquitous issue is how to aggregate the instance labels to generate a bag label. This problem also takes place in MIPL and is more difficult due to the ambiguous multi-class classification. A feasible solution is to set the bag label with the class label corresponding to the maximum value among the class probabilities of all instances in the bag. Let $\widetilde{\bf{\Theta}}_i = [\bs{\theta}_i^1, \bs{\theta}_i^2,\cdots,\bs{\theta}_i^{z_i}]^\top \in \bb{R}^{z_i \times \tilde{q}}$ ($\bs{\theta}_i^j = [\theta_i^1, \theta_i^2,\cdots, \theta_i^q, \theta_i^{neg}]^\top$ for $j=1,2,\cdots,z_i$) denote the class probabilities of the multi-instance bag $\bs{X}_i$ among $\tilde{q}$ outputs. We truncate $\widetilde{\bf{\Theta}}_i$ to yield ${\bf{\Theta}}_i = [\hat{\bs{\theta}}_i^1, \hat{\bs{\theta}}_i^2,\cdots,\hat{\bs{\theta}}_i^{z_i}]^\top \in \bb{R}^{z_i \times q}$ among $q$ outputs, where $\hat{\bs{\theta}}_i^j = [\theta_i^1, \theta_i^2,\cdots, \theta_i^q]^\top$. Consequently, the aggregated bag label is as follows: 
\begin{equation}
\label{eq:bag_label_1}
	Y_{i} = \psi ({\bf{\Theta}}_i),
\end{equation}
where the goal of $\psi({\bf{\Theta}}_i)$ is to query the column index corresponding to the maximum value in ${\bf{\Theta}}_i$. The plate diagram of {M\scriptsize{IPL}}G{\scriptsize{P}} is illustrated in Figure \ref{fig:diagram}, where the grey circles represent the observed variables, i.e., features and bag labels, and the white circles represent the latent variables.

In the training phase, the model parameters $\Phi$ are learned by minimizing the negative log marginal likelihood:
\begin{equation}
\label{eq:training}
	\al{L} = - \log P(\dot{\bf{Y}} \mid {\bf{X}}, \Phi) \propto  \log \left|{\bf{K}}\right| + \dot{\bf{Y}}^{\top} {\bf{K}}^{-1} \dot{\bf{Y}},
\end{equation}
and the derivative is as follows:
\begin{equation}
\label{eq:derivative}
	\frac{\partial \al{L}}{\partial \Phi} \propto \operatorname{Tr}\left({\bf{K}}^{-1} \frac{\partial {\bf{K}}}{\partial \Phi} \right) - \dot{\bf{Y}}^{\top} {\bf{K}} \frac{\partial {\bf{K}}^{-1}}{\partial \Phi} {\bf{K}} \dot{\bf{Y}},
\end{equation}
where $\operatorname{Tr}(\cdot)$ is the trace operation.
In a naive GP, the Cholesky decomposition of ${\bf{K}^{-1}}$ requires $\al{O}(\tilde{q}n^3)$ computations. However, we employ an efficient algorithm with GPU accelerations and reduce the complexity to $\al{O}(\tilde{q}n^2)$.
Moreover, a preconditioner is adopted to further accelerate the computation \cite{WangPGTWW19}.

Given an unseen multi-instance bag $\bs{X}_{*} = [\bs{x}^1, \bs{x}^2,\cdots,$ $ \bs{x}^{z_*}]$ with $z_*$ instances, the GP model generates the predictive distribution $P({\bf{F}}^* | {\bf{X}}, \dot{{\bf{Y}}}, \bs{x}^{i_*})$ ($i_* = 1,2,\cdots,$ $z_*$) for the corresponding latent variables ${\bf{F}^*} = [f^{*1}, f^{*2},\cdots, f^{*\tilde{q}}]^\top$.	
The Dirichlet posterior of the class label is procured from the predictive distribution using the Monte Carlo method so as to calculate the expectation of class probability:
\begin{equation}
\label{eq:pred}
	\bb{E}[\theta_{i_*}^c \mid {\bf{X}}, \dot{{\bf{Y}}}, \bs{x}^{i_*} ] = \int_{\bf{F}^*}  \frac{\exp \left(f^{*c}\right(\bs{x}^{i_*}))}{\sum_{j=1}^{\tilde{q}} \exp \left(f^{*j}(\bs{x}^{i_*}) \right)}  P(f^{*c}(\bs{x}^{i_*}) \mid {\bf{X}}, \dot{{\bf{Y}}}, \bs{x}^{i_*}),
\end{equation}
where $P(f^{*c}(\bs{x}^{i_*}) \mid {\bf{X}}, \dot{{\bf{Y}}}, \bs{x}^{i_*})$ is the predictive distribution for $f^{*c}(\cdot, \cdot)$.
We write the class probability of $\bs{x}^{i_*}$ as $\bs{\theta}^{i_*}_* = [\theta_{i_*}^1, \theta_{i_*}^2,\cdots, \theta_{i_*}^{q}, \theta_{i_*}^{neg}]$, and let $\widetilde{\bf{\Theta}}_* = [\bs{\theta}_*^1, \bs{\theta}_*^2,\cdots,\bs{\theta}_*^{z_*}]^\top \in \bb{R}^{z_* \times \tilde{q}}$ denote the class probabilities of all instances in the test bag $\bs{X}_{*}$.
Then, we consider the label space $\al{Y}$ instead of $\widetilde{\al{Y}}$, i.e., remove the probabilities corresponding to class $l_{neg}$ and get ${\bf{\Theta}}_* \in \bb{R}^{z_* \times q}$ with $z_*$ instances and $q$ class labels. 
Finally, a bag label is predicted by:
\begin{equation}
\label{eq:bag_label}
	Y_* = \psi ({\bf{\Theta}}_*).
\end{equation}

Algorithm \ref{alg:algorithm} summarizes the complete procedure of {M\scriptsize{IPL}}G{\scriptsize{P}}. First, the algorithm propagates the augmented candidate label set of each bag to all instances in the bag (Steps 1-7). After initializing the shape parameter of the Dirichlet distribution (Step 8), the Gaussian processes model is induced based on the transformed labels (Steps 9-21). Last, the label of an unseen multi-instance bag is returned by querying the predicted class probabilities (Step 22).

\begin{algorithm}[!h]	
\caption{$Y_* = \text{M}{\scriptsize{\text{IPL}}}\text{G}{\scriptsize{\text{P}}} (\al{D}, \alpha_{\epsilon}, T, \bs{X}_{*})$ }
\label{alg:algorithm}
\textbf{Inputs}: \\
$\al{D}$ : the multi-instance partial-label training set $\{(\bs{X}_i,\bs{y}_i)\mid 1 \le i \le m \}$ ($\bs{X}_i=[\bs{x}_i^1,\bs{x}_i^2, \cdots, \\~~~~~~~\bs{x}_i^{z_i}]^\top$, $\bs{x}_i^{z_i} \in \al{X}$, $\bs{X}_i \subseteq   \al{X}$, $\al{X}=\bb{R}^d$, 
$\bs{y}_i = [y_i^1,y_i^2, \cdots, y_i^q]^\top$, $\bs{y}_i \subseteq \al{Y}$, $\al{Y}=\{l_1, l_2, \cdots, l_q\}$)\\
$\alpha_{\epsilon}$ : the Dirichlet prior  \\
\textit{T} : the number of iterations \\
$\bs{X}_{*}$: the unseen multi-instance bags with $z_*$ instances\\
\textbf{Outputs}: \\
$\bs{Y}_{*}$ : the predicted bag label for $\bs{X}_{*}$ \\
\textbf{Process}: 
\begin{algorithmic}[1]  
\STATE Augment $\al{Y}$ to $\widetilde{\al{Y}}=\{l_1, l_2,\cdots, l_q, l_{neg}\}$ 
\FOR{$i=1$ to $m$}
	\STATE Augment $\bs{y}_i$ to $\widetilde{\bs{y}}_i = [y_i^1,y_i^2,\cdots, y_i^q, y_i^{neg}]^\top$   
	\FOR{$j=1$ to $z_i$}
		\STATE Propagate $\widetilde{\bs{y}}_i$ to be the label of each instance $\bs{x}_i^j$
	\ENDFOR
\ENDFOR
\STATE Initialize $\bs{\alpha}_i$ as defined in Eq. (\ref{eq:init_weight}) 
\FOR{$t=1$ to $T$}
	\FOR{$i=1$ to $m$}
		\FOR{$j=1$ to $z_i$}
			\STATE Model the likelihood $p(\bs{\widetilde{y}}_i \mid \bs{\alpha}_i) = \text{Cat}(\bs{\theta}_i)$,  $\bs{\theta}_i \sim \text{Dir}({\bs{\alpha}_i})$ according to Eq. (\ref{eq:likelihood})
			\STATE Generate Dirichlet samples from the Gamma distribution according to Eq. (\ref{eq:gamma})
			\STATE Derive the continuous candidate label set $\dot{\bs{y}}_i$ as stated by Eq. (\ref{eq:new_label})
		\ENDFOR
	\ENDFOR
	\STATE Calculate $\al{L}$ according to Eq. (\ref{eq:training})
	\STATE Calculate gradient $\frac{\partial \al{L}}{\partial \Phi}$ as defined in Eq. (\ref{eq:derivative})
	\STATE Update $\Phi$ by the optimizer
	\STATE Update $\bs{\alpha}_i$ according to Eq. (\ref{eq:update_weight})
\ENDFOR
\STATE Return $Y_*$ according to Eq. (\ref{eq:bag_label})
\end{algorithmic}
\end{algorithm}

\section{Experiments}
\label{sec:experiments}
\subsection{Experimental Setup}
\subsubsection{Datasets}
To the best of our knowledge, there are no ready-made datasets for solving the MIPL problems. To overcome this limitation, we synthesize five MIPL datasets stemming from relevant literature \cite{lecun1998gradient, han1708, Lang95, briggs2012rank, SettlesCR07}, i.e., MNIST-{\scriptsize{MIPL}}, FMNIST-{\scriptsize{MIPL}}, Newsgroups-{\scriptsize{MIPL}}, Birdsong-{\scriptsize{MIPL}}, and SIVAL-{\scriptsize{MIPL}}, from domains of image, text, and biology to compare {M\scriptsize{IPL}}G{\scriptsize{P}} with other algorithms. 

The characteristics of the synthetic datasets are summarized in Table \ref{tab:datasets}. We use \emph{\#bags}, \emph{\#ins}, \emph{\#max}, \emph{\#min}, \emph{\#dims}, \emph{\#l-o}, \emph{\#l-t}, \emph{\#l-r}, and \emph{percentage} to denote the number of bags, number of instances, maximum number of instances in a bag, minimum number of instances in a bag, dimension of each instance, number of targeted class labels in the corresponding literature, number of targeted class labels in MIPL, number of reserved class labels, and percentage of the number of positive instances in each dataset.  
\begin{table}[t] 
\centering \scriptsize
  \caption{Characteristics of the MIPL Datasets.}
  \label{tab:datasets}
\begin{tabular}{lccccccccccc}
\hline \hline
	Dataset				& \#bags	& \#ins	&\#max	&\#min	& \#dims	& \#l-o	& \#l-t 	& \#l-r	& percentage	& domain		\\ 	\hline
	MNIST-{\tiny{MIPL}}		& 500	& 20664	& 48		& 35		& 784	& 10		& 5		& 5		& $8.0\%$		& image		\\ 
	FMNIST-{\tiny{MIPL}}	& 500	& 20810	& 48		& 36		& 784	& 10		& 5		& 5		& $8.0\%$		& image		\\ 
	Newsgroups-{\tiny{MIPL}}	& 1000	& 43122	& 86		& 11		& 200	& 20		& 10		& 10		& $8.0\%$		& text		\\ 
	Birdsong-{\tiny{MIPL}}	& 1300	& 48425	& 76		& 25		& 38		& 13		& 13		& 1		& $8.3\%$		& biology		\\ 
	SIVAL-{\tiny{MIPL}}		& 1500	& 47414	& 32		& 31		& 30		& 25		& 25		& -- 		& $25.6\%$	& image		\\ 
	  \hline \hline
  \end{tabular}
\end{table}

To synthesize the multi-instance bag with a candidate label set, we take positive instances of the corresponding label from the targeted class labels and negative instances in the whole reserved class labels. Furthermore, we sample the false positive labels from the target classes without replacement. For comprehensive performance evaluation, the number of false positive labels depends on the controlling parameter $r$ ($|\bs{y}_i| = r + 1$). Note that we sample all instances and generate false positive labels randomly and uniformly. 

Take MNIST-{\scriptsize{MIPL}}  for example,  the number of MNIST is 10 in most classification problems. To obtain a multi-instance bag in the MIPL, we extract $\{0,2,4,6,8\}$ as five target classes for providing the positive instances according to the corresponding class and draw all negative ones from the reserved classes $\{1,3,5,7,9\}$ randomly. Then, we append any $r$ false positive labels from the target classes to the candidate label set of the multi-instance bag. More detailed information on the MIPL datasets is provided in Appendix A. 

\subsubsection{Comparative Algorithms}
{M\scriptsize{IPL}}G{\scriptsize{P}} is compared against four well-established multi-instances learning algorithms including three Gaussian processes-based algorithms {V\scriptsize{WSGP}} \cite{KandemirHDRLH16}, {V\scriptsize{GPMIL}} \cite{HauBmannHK17}, and {L\scriptsize{M-VGPMIL}} \cite{HauBmannHK17}, as well as variational autoencoder based algorithm {M\scriptsize{IVAE}} \cite{Weijia21}. 
In addition, we employ six partial-label learning algorithms containing two averaging-based algorithms {P\scriptsize{L-kNN}} \cite{hullermeier2006learning} and {C\scriptsize{LPL}} \cite{cour2011learning}, two identification-based algorithms {L\scriptsize{SB-CMM}} \cite{liu2012conditional} and {S\scriptsize{URE}} \cite{feng2019partial}, graph matching based disambiguation algorithm {G\scriptsize{M-PLL}} \cite{lyu2019gm}, and feature-aware disambiguation algorithm {P\scriptsize{L-AGGD}} \cite{wang2021adaptive}. The parameter configurations of each algorithm are suggested in the respective literature.

To verify the effectiveness of the Dirichlet disambiguation and the label augmentation, we evaluate two variants of {M\scriptsize{IPL}}G{\scriptsize{P}}, i.e., {M\scriptsize{IPL}}G{\scriptsize{P}}-uniform and {M\scriptsize{IPL}}G{\scriptsize{P}}-naive. The former utilizes the uniform weights as defined in Eq. (\ref{eq:init_weight}) throughout the iterations, and the latter handles the candidate label sets only in the original label space $\al{Y}$.

\subsubsection{Implementation} 
We implement {M\scriptsize{IPL}}G{\scriptsize{P}} using GPyTorch, which is a modular Gaussian process library in PyTorch \cite{GardnerPWBW18}.
For {M\scriptsize{IPL}}G{\scriptsize{P}} and its variants, we use the Adam optimizer \cite{KingmaB14} with $\beta_1=0.9$ and $\beta_2=0.999$. The initial learning rate is $0.1$ which is decayed via a cosine annealing method \cite{LoshchilovH17}. We set the number of iterations to $500$ for MNIST-{\scriptsize{MIPL}}  and FMNIST-{\scriptsize{MIPL}}  datasets and $50$ for the remaining three datasets.
For Newsgroups-{\scriptsize{MIPL}}  dataset, the smoothness parameter $\nu$ in Mat{\'{e}}rn kernel is $0.5$, while $\nu=2.5$ for the rest. We uniform that $\ell$ in Mat{\'{e}}rn kernel is equal to $1$, $\alpha_{\epsilon}=0.0001$, the size of the preconditioner is $100$, and the number of Monte Carlo sampling points is $512$ for all datasets. 
We perform ten runs of $50\%/50\%$ random train/test splits on all datasets, and record the mean accuracies and standard deviations for each algorithm.
Specifically, we conduct the pairwise t-test at a $0.05$ significance level based on the results of ten runs. Experiments are mainly conducted with two Nvidia Tesla V100 GPUs.

\subsection{Experimental Results}
Since multi-instance learning and partial-label learning are both special cases of MIPL, we carry out experiments with multi-instance learning and partial-label learning algorithms using different versions of degenerated MIPL datasets. However, the experiments with {M\scriptsize{IPL}}G{\scriptsize{P}} and its variants are executed on the standard MIPL datasets.

\subsubsection{Comparison with Partial Learning Algorithms}
Existing partial-label learning algorithms are incapable of handling the multi-instance bag. Therefore, an aggregated feature representation of the bag is a prerequisite for tackling the MIPL problems with partial-label learning algorithms. In multi-instance learning, the idea of the embedded-space paradigm is to explicitly distill the whole bag by defining a mapping function from the bag to a feature vector.  
Inspired by the idea, we utilize two schemes to map the bag to a holistic feature vector, respectively.
\begin{enumerate}
\setlength{\itemsep}{-10pt}
\setlength{\parsep}{0pt}
\setlength{\parskip}{0pt}
	\item[•] \textbf{Mean} scheme: For each bag, the average value of all instances in the corresponding feature dimension is calculated as the final feature value in that dimension. The dimension of the holistic feature vector is the same as that of each instance.\\
	\item[•] \textbf{MaxMin} scheme: We choose the maximum values of all instances in each feature dimension and concatenate them with the minimum values of all instances in each feature dimension. Finally, the holistic feature representation of a bag is distilled with the length of $2d$.
\end{enumerate}

\begin{table}[!t]
  \centering \scriptsize
   \caption{Classification accuracy (mean$\pm$std) of each comparing algorithm in terms of the different number of false positive candidate labels [$r \in \{1,2,3\}$]. $\bullet/\circ$ indicates whether the performance of {M\scriptsize{IPL}}G{\scriptsize{P}} is statistically superior/inferior to the comparing algorithm on each dataset (pairwise t-test at $0.05$ significance level).}  
  \label{tab:pll}
    \begin{tabular}{r|c|ccccc}
    \hline  \hline
    \multicolumn{1}{l|}{Algorithm} & $r$     & MNIST-{\tiny{MIPL}} & FMNIST-{\tiny{MIPL}} & Newsgroups-{\tiny{MIPL}} & Birdsong-{\tiny{MIPL}} & SIVAL-{\tiny{MIPL}} \\
    \hline
    \multicolumn{1}{l|}{\multirow{3}[2]{*}{{M\scriptsize{IPL}}G{\scriptsize{P}}}} 
    	& 1     & 0.921$\pm$0.018 & 0.806$\pm$0.031 & 0.432$\pm$0.018 & 0.628$\pm$0.012 & 0.599$\pm$0.020 \\
       	& 2     & 0.712$\pm$0.045 & 0.778$\pm$0.042 & 0.424$\pm$0.019 & 0.589$\pm$0.020 & 0.535$\pm$0.020 \\
       	& 3     & 0.521$\pm$0.084 & 0.592$\pm$0.076 & 0.373$\pm$0.023 & 0.538$\pm$0.014 & 0.497$\pm$0.023 \\
    \hline
    \multicolumn{1}{l|}{\multirow{3}[2]{*}{{M\scriptsize{IPL}}G{\scriptsize{P}}-uniform}} 
    	& 1     & 0.834$\pm$0.023$\bullet$ & 0.778$\pm$0.031$\bullet$ & 0.417$\pm$0.019$\bullet$ & 0.623$\pm$0.013$\bullet$ & 0.595$\pm$0.023 \\
       	& 2     & 0.531$\pm$0.070$\bullet$ & 0.746$\pm$0.042$\bullet$ & 0.401$\pm$0.025$\bullet$ & 0.581$\pm$0.022$\bullet$ & 0.530$\pm$0.019$\bullet$ \\
       	& 3     & 0.206$\pm$0.009$\bullet$ & 0.226$\pm$0.039$\bullet$ & 0.365$\pm$0.013 	     & 0.526$\pm$0.016$\bullet$ & 0.489$\pm$0.026$\bullet$ \\
    \hline
    \multicolumn{1}{l|}{\multirow{3}[2]{*}{{M\scriptsize{IPL}}G{\scriptsize{P}}-naive}} 
    	& 1     & 0.522$\pm$0.025$\bullet$ & 0.570$\pm$0.016$\bullet$ & 0.422$\pm$0.019$\bullet$ & 0.551$\pm$0.010$\bullet$ & 0.585$\pm$0.019$\bullet$ \\
       	& 2     & 0.438$\pm$0.049$\bullet$ & 0.468$\pm$0.065$\bullet$ & 0.407$\pm$0.025$\bullet$ & 0.511$\pm$0.026$\bullet$ & 0.523$\pm$0.018$\bullet$ \\
       	& 3     & 0.309$\pm$0.072$\bullet$ & 0.258$\pm$0.045$\bullet$ & 0.373$\pm$0.017 	     & 0.464$\pm$0.019$\bullet$ & 0.480$\pm$0.021$\bullet$ \\
    \hline
    \multicolumn{7}{c}{Mean} \\
    \hline
    \multicolumn{1}{l|}{\multirow{3}[2]{*}{{P\scriptsize{L-kNN}}}} 
    	& 1     & 0.397$\pm$0.021$\bullet$ & 0.419$\pm$0.032$\bullet$ & 0.133$\pm$0.009$\bullet$ & 0.213$\pm$0.011$\bullet$ & 0.155$\pm$0.009$\bullet$ \\
       	& 2     & 0.337$\pm$0.020$\bullet$ & 0.360$\pm$0.030$\bullet$ & 0.148$\pm$0.006$\bullet$ & 0.197$\pm$0.012$\bullet$ & 0.138$\pm$0.008$\bullet$ \\
       	& 3     & 0.284$\pm$0.023$\bullet$ & 0.264$\pm$0.032$\bullet$ & 0.142$\pm$0.010$\bullet$ & 0.182$\pm$0.009$\bullet$ & 0.123$\pm$0.009$\bullet$ \\
    \hline
    \multicolumn{1}{l|}{\multirow{3}[2]{*}{{C\scriptsize{LPL}}}} 
    	& 1     & 0.644$\pm$0.023$\bullet$ & 0.734$\pm$0.031$\bullet$ & 0.131$\pm$0.027$\bullet$ & 0.330$\pm$0.013$\bullet$ & 0.239$\pm$0.009$\bullet$ \\
       	& 2     & 0.528$\pm$0.033$\bullet$ & 0.671$\pm$0.023$\bullet$ & 0.112$\pm$0.015$\bullet$ & 0.295$\pm$0.012$\bullet$ & 0.221$\pm$0.012$\bullet$ \\
       	& 3     & 0.377$\pm$0.042$\bullet$ & 0.524$\pm$0.046 		  & 0.111$\pm$0.012$\bullet$ & 0.282$\pm$0.011$\bullet$ & 0.200$\pm$0.017$\bullet$ \\
    \hline
    \multicolumn{1}{l|}{\multirow{3}[2]{*}{{L\scriptsize{SB-CMM}}}} 
    	& 1     & 0.631$\pm$0.045$\bullet$ & 0.709$\pm$0.025$\bullet$ & 0.100$\pm$0.000$\bullet$ & 0.260$\pm$0.013$\bullet$ & 0.144$\pm$0.012$\bullet$ \\
       	& 2     & 0.416$\pm$0.047$\bullet$ & 0.560$\pm$0.059$\bullet$ & 0.100$\pm$0.000$\bullet$ & 0.242$\pm$0.013$\bullet$ & 0.116$\pm$0.014$\bullet$ \\
       	& 3     & 0.277$\pm$0.038$\bullet$ & 0.295$\pm$0.032$\bullet$ & 0.100$\pm$0.000$\bullet$ & 0.218$\pm$0.011$\bullet$ & 0.095$\pm$0.015$\bullet$ \\
    \hline
    \multicolumn{1}{l|}{\multirow{3}[2]{*}{{S\scriptsize{URE}}}} 
    	& 1     & 0.666$\pm$0.027$\bullet$ & 0.753$\pm$0.019$\bullet$ & 0.358$\pm$0.019$\bullet$ & 0.345$\pm$0.008$\bullet$ & 0.313$\pm$0.022$\bullet$ \\
       	& 2     & 0.512$\pm$0.031$\bullet$ & 0.685$\pm$0.013$\bullet$ & 0.300$\pm$0.013$\bullet$ & 0.319$\pm$0.008$\bullet$ & 0.284$\pm$0.019$\bullet$ \\
       	& 3     & 0.344$\pm$0.075$\bullet$ & 0.441$\pm$0.063$\bullet$ & 0.251$\pm$0.017$\bullet$ & 0.308$\pm$0.013$\bullet$ & 0.256$\pm$0.013$\bullet$ \\
    \hline
    \multicolumn{1}{l|}{\multirow{3}[2]{*}{{G\scriptsize{M-PLL}}}} 
    	& 1     & 0.248$\pm$0.014$\bullet$ & 0.268$\pm$0.030$\bullet$ & 0.183$\pm$0.012$\bullet$ & 0.146$\pm$0.018$\bullet$ & 0.175$\pm$0.013$\bullet$ \\
       	& 2     & 0.260$\pm$0.022$\bullet$ & 0.251$\pm$0.019$\bullet$ & 0.180$\pm$0.015$\bullet$ & 0.106$\pm$0.012$\bullet$ & 0.160$\pm$0.014$\bullet$ \\
       	& 3     & 0.235$\pm$0.028$\bullet$ & 0.246$\pm$0.021$\bullet$ & 0.157$\pm$0.015$\bullet$ & 0.092$\pm$0.012$\bullet$ & 0.136$\pm$0.013$\bullet$ \\
    \hline
    \multicolumn{1}{l|}{\multirow{3}[2]{*}{{P\scriptsize{L-AGGD}}}} 
    	& 1     & 0.643$\pm$0.021$\bullet$ & 0.714$\pm$0.017$\bullet$ & 0.325$\pm$0.009$\bullet$ & 0.332$\pm$0.010$\bullet$ & 0.312$\pm$0.023$\bullet$ \\
       	& 2     & 0.535$\pm$0.034$\bullet$ & 0.642$\pm$0.026$\bullet$ & 0.256$\pm$0.015$\bullet$ & 0.304$\pm$0.013$\bullet$ & 0.277$\pm$0.019$\bullet$ \\
       	& 3     & 0.363$\pm$0.039$\bullet$ & 0.429$\pm$0.040$\bullet$ & 0.213$\pm$0.015$\bullet$ & 0.292$\pm$0.015$\bullet$ & 0.244$\pm$0.011$\bullet$ \\
    \hline
    \multicolumn{7}{c}{MaxMin} \\
    \hline
    \multicolumn{1}{l|}{\multirow{3}[2]{*}{{P\scriptsize{L-kNN}}}} 
    	& 1     & 0.388$\pm$0.023$\bullet$ & 0.309$\pm$0.029$\bullet$ & 0.119$\pm$0.004$\bullet$ & 0.274$\pm$0.009$\bullet$ & 0.177$\pm$0.007$\bullet$ \\
       	& 2     & 0.330$\pm$0.016$\bullet$ & 0.288$\pm$0.019$\bullet$ & 0.135$\pm$0.008$\bullet$ & 0.270$\pm$0.006$\bullet$ & 0.153$\pm$0.009$\bullet$ \\
       	& 3     & 0.266$\pm$0.025$\bullet$ & 0.239$\pm$0.021$\bullet$ & 0.138$\pm$0.007$\bullet$ & 0.250$\pm$0.011$\bullet$ & 0.136$\pm$0.011$\bullet$ \\
    \hline
    \multicolumn{1}{l|}{\multirow{3}[2]{*}{{C\scriptsize{LPL}}}} 
    	& 1     & 0.481$\pm$0.020$\bullet$ & 0.364$\pm$0.026$\bullet$ & 0.246$\pm$0.009$\bullet$ & 0.361$\pm$0.016$\bullet$ & 0.266$\pm$0.011$\bullet$ \\
       	& 2     & 0.396$\pm$0.028$\bullet$ & 0.332$\pm$0.021$\bullet$ & 0.200$\pm$0.009$\bullet$ & 0.328$\pm$0.014$\bullet$ & 0.223$\pm$0.010$\bullet$ \\
       	& 3     & 0.334$\pm$0.039$\bullet$ & 0.332$\pm$0.028$\bullet$ & 0.166$\pm$0.018$\bullet$ & 0.300$\pm$0.015$\bullet$ & 0.199$\pm$0.014$\bullet$ \\
    \hline
    \multicolumn{1}{l|}{\multirow{3}[2]{*}{{L\scriptsize{SB-CMM}}}} 
   		& 1     & 0.372$\pm$0.099$\bullet$ & 0.238$\pm$0.073$\bullet$ & 0.221$\pm$0.018$\bullet$ & 0.319$\pm$0.011$\bullet$ & 0.248$\pm$0.015$\bullet$ \\
       	& 2     & 0.324$\pm$0.038$\bullet$ & 0.284$\pm$0.039$\bullet$ & 0.146$\pm$0.039$\bullet$ & 0.292$\pm$0.014$\bullet$ & 0.200$\pm$0.017$\bullet$ \\
       	& 3     & 0.220$\pm$0.017$\bullet$ & 0.210$\pm$0.017$\bullet$ & 0.113$\pm$0.021$\bullet$ & 0.272$\pm$0.020$\bullet$ & 0.157$\pm$0.017$\bullet$ \\
    \hline
    \multicolumn{1}{l|}{\multirow{3}[2]{*}{{S\scriptsize{URE}}}} 
    	& 1     & 0.528$\pm$0.021$\bullet$ & 0.404$\pm$0.023$\bullet$ & 0.316$\pm$0.015$\bullet$ & 0.381$\pm$0.013$\bullet$ & 0.372$\pm$0.022$\bullet$ \\
       	& 2     & 0.415$\pm$0.028$\bullet$ & 0.351$\pm$0.025$\bullet$ & 0.274$\pm$0.013$\bullet$ & 0.371$\pm$0.015$\bullet$ & 0.324$\pm$0.013$\bullet$ \\
       	& 3     & 0.321$\pm$0.030$\bullet$ & 0.304$\pm$0.029$\bullet$ & 0.245$\pm$0.014$\bullet$ & 0.341$\pm$0.014$\bullet$ & 0.288$\pm$0.011$\bullet$ \\
    \hline
    \multicolumn{1}{l|}{\multirow{3}[2]{*}{{G\scriptsize{M-PLL}}}} 
    	& 1     & 0.386$\pm$0.024$\bullet$ & 0.195$\pm$0.016$\bullet$ & 0.209$\pm$0.020$\bullet$ & 0.180$\pm$0.019$\bullet$ & 0.143$\pm$0.013$\bullet$ \\
       	& 2     & 0.346$\pm$0.030$\bullet$ & 0.225$\pm$0.019$\bullet$ & 0.181$\pm$0.018$\bullet$ & 0.139$\pm$0.023$\bullet$ & 0.121$\pm$0.014$\bullet$ \\
       	& 3     & 0.294$\pm$0.024$\bullet$ & 0.221$\pm$0.013$\bullet$ & 0.163$\pm$0.019$\bullet$ & 0.121$\pm$0.020$\bullet$ & 0.104$\pm$0.012$\bullet$ \\
    \hline
    \multicolumn{1}{l|}{\multirow{3}[2]{*}{{P\scriptsize{L-AGGD}}}} 
    	& 1     & 0.514$\pm$0.024$\bullet$ & 0.392$\pm$0.016$\bullet$ & 0.289$\pm$0.014$\bullet$ & 0.370$\pm$0.013$\bullet$ & 0.361$\pm$0.021$\bullet$ \\
       	& 2     & 0.428$\pm$0.035$\bullet$ & 0.346$\pm$0.019$\bullet$ & 0.249$\pm$0.013$\bullet$ & 0.353$\pm$0.014$\bullet$ & 0.310$\pm$0.013$\bullet$ \\
       	& 3     & 0.333$\pm$0.039$\bullet$ & 0.324$\pm$0.025$\bullet$ & 0.212$\pm$0.011$\bullet$ & 0.328$\pm$0.015$\bullet$ & 0.277$\pm$0.013$\bullet$ \\
    \hline  \hline
    \end{tabular}
\end{table}

\begin{table}[!t]
  \centering \footnotesize
    \begin{tabular}{ccccccc|c}
    \hline \hline
          & \multicolumn{6}{l|}{{M\scriptsize{IPL}}G{\scriptsize{P}} against}  & \multirow{2}[2]{*}{\textbf{In total}} \\ \cdashline{2-7}[2pt/1pt]
          & {P\scriptsize{L-kNN}} & {C\scriptsize{LPL}}  & {L\scriptsize{SB-CMM}} & {S\scriptsize{URE}}  & {G\scriptsize{M-PLL}} & {P\scriptsize{L-AGGD}} &  \\
    \hline
    $r=1$   & 10/0/0 & 10/0/0 & 10/0/0 & 10/0/0 & 10/0/0 & 10/0/0 & 60/0/0 \\
    $r=2$   & 10/0/0 & 10/0/0 & 10/0/0 & 10/0/0 & 10/0/0 & 10/0/0 & 60/0/0 \\
    $r=3$   & 10/0/0 & 9/1/0  & 10/0/0  & 10/0/0 & 10/0/0 & 10/0/0 & 59/1/0 \\
    \hline
    \textbf{In total} & 30/0/0 & 29/1/0 & 30/0/0 & 30/0/0 & 30/0/0 & 30/0/0 & \textbf{179/1/0} \\
    \hline \hline
    \end{tabular}
  \caption{Win/tie/loss counts on the classification performance of {M\scriptsize{IPL}}G{\scriptsize{P}} against the comparing PLL algorithms.}
  \label{tab:win_pll}
\end{table}

The classification results with the varying number of false positive labels $r$ are reported in Table \ref{tab:pll}, and Table \ref{tab:win_pll} summarizes the win/tie/loss counts between {M\scriptsize{IPL}}G{\scriptsize{P}} and each comparing algorithm. {M\scriptsize{IPL}}G{\scriptsize{P}} achieves superior or competitive performance against the comparing algorithms. Out of the $180$ statistical tests, we yield the following observations:
\begin{enumerate}
\setlength{\itemsep}{-10pt}
\setlength{\parsep}{0pt}
\setlength{\parskip}{0pt}
	\item[•] {M\scriptsize{IPL}}G{\scriptsize{P}} is statistically superior to the comparing partial-label learning algorithms in $99.444\%$ of the cases. \\
	\item[•] Compared to {M\scriptsize{IPL}}G{\scriptsize{P}}-uniform and {M\scriptsize{IPL}}G{\scriptsize{P}}-naive, {M\scriptsize{IPL}}G{\scriptsize{P}} achieves statistically favorable performance in $86.667\%$ and $93.333\%$ of the cases, respectively. \\
	\item[•] Regardless of the Mean scheme or MaxMin scheme, {M\scriptsize{IPL}}G{\scriptsize{P}} consistently outperforms the comparing partial-label learning algorithms by a notable margin, e.g., more than $5.3$ percent, in almost all cases. \\
	\item[•] In most cases, {M\scriptsize{IPL}}G{\scriptsize{P}}-uniform is superior to {M\scriptsize{IPL}}G{\scriptsize{P}}-naive, which means that the label augmentation strategy plays an important role in {M\scriptsize{IPL}}G{\scriptsize{P}}. As the average accuracy of {M\scriptsize{IPL}}G{\scriptsize{P}}-uniform decreases faster than that of {M\scriptsize{IPL}}G{\scriptsize{P}}-naive as the number of false positive labels increases, the results demonstrate that the Dirichlet disambiguation is indispensable especially when there are a lot of false positive labels.
\end{enumerate}

\subsubsection{Comparison with Multi-Instance Learning Algorithms}
Most of the existing multi-instance learning algorithms are only designed to solve binary classification problems, and thus are not directly applicable to the MIPL problems.

To make multi-instance learning algorithms fit the MIPL problems, we employ the \emph{One vs. Rest} (\emph{OvR}) decomposition strategy. Specifically, given a multi-instance bag $\bs{X}_i$ associated with a candidate label set $\bs{y}_i$, we assign each label in the candidate label set to the bag in turn and yield $|\bs{y}_i|$ multi-instance bags with a single bag-level label. For $c=1,2,\cdots,q$, we recompose the label $c$ to $1$, i.e., positive, and other labels to $0$, i.e., negative. After recomposing all multi-instance bags for the label $c$, we train and test the $c$-th classifier. For an unseen multi-instance bag, we can obtain $q$ predictions from the $q$ classifiers. If only one of the predictions is positive, the corresponding class label of the positive prediction is regarded as the classification result of the bag. 
If the number of positive predictions among the $q$ predictions is greater than one, the class label corresponding to the classifier with the largest prediction confidence is selected as the classification result of the bag. If the predictions of $q$ classifiers all are negative, the classification result is the class label with the lowest prediction confidence.

\begin{table}[!b]
 \caption{Classification accuracy (mean$\pm$std) of each comparing algorithm (with one false positive candidate label [$r = 1$]). $\bullet/\circ$ indicates whether the performance of {M\scriptsize{IPL}}G{\scriptsize{P}} is statistically superior/inferior to the comparing algorithm on each dataset (pairwise t-test at $0.05$ significance level).}
  \label{tab:mll}%
  \centering \footnotesize
    \begin{tabular}{lccccc}
    \hline \hline
    Algorithm 					& MNIST-{\scriptsize{MIPL}} 	& FMNIST-{\scriptsize{MIPL}} 	& Newsgroups-{\scriptsize{MIPL}} 	& Birdsong-{\scriptsize{MIPL}}	& SIVAL-{\scriptsize{MIPL}} \\
    \hline
    {M\scriptsize{IPL}}G{\scriptsize{P}}& 0.921$\pm$0.018 			& 0.806$\pm$0.031 			& 0.432$\pm$0.018 				& 0.628$\pm$0.012 			& 0.599$\pm$0.020 \\
    {V\scriptsize{WSGP}} 			& 0.402$\pm$0.026$\bullet$	& 0.422$\pm$0.028$\bullet$ 	& 0.098$\pm$0.013$\bullet$ 		& 0.250$\pm$0.047$\bullet$ 	& 0.050$\pm$0.009$\bullet$ \\
    {V\scriptsize{GPMIL}} 			& 0.469$\pm$0.047$\bullet$ 	& 0.455$\pm$0.034$\bullet$ 	& 0.097$\pm$0.010$\bullet$		& 0.080$\pm$0.034$\bullet$ 	& 0.041$\pm$0.006$\bullet$ \\
    {L\scriptsize{M-VGPMIL}} 		& 0.471$\pm$0.021$\bullet$ 	& 0.486$\pm$0.036$\bullet$ 	& 0.101$\pm$0.008$\bullet$ 		& 0.081$\pm$0.042$\bullet$ 	& 0.045$\pm$0.008$\bullet$ \\
    {M\scriptsize{IVAE}} 			& 0.793$\pm$0.019$\bullet$	& 0.638$\pm$0.213 			& 0.135$\pm$0.245$\bullet$ 		& 0.067$\pm$0.091$\bullet$ 	& 0.068$\pm$0.119$\bullet$ \\
    \hline \hline
    \end{tabular}
\end{table}
The computational cost of multi-instance learning algorithms rises with the increase of false positive labels, and the classification accuracy decreases accordingly.
We present the classification accuracy of multi-instance learning algorithms with one false positive label in Table \ref{tab:mll}, which reveals that:
\begin{enumerate}
\setlength{\itemsep}{-10pt}
\setlength{\parsep}{0pt}
\setlength{\parskip}{0pt}
	\item[•] {M\scriptsize{IPL}}G{\scriptsize{P}} achieves significantly better performances against the comparing multi-instance learning algorithms in almost all cases. \\
	\item[•] Due to the noisy bag-level labels in the degenerated datasets, the comparing multi-instance learning algorithms can learn well in multi-instance learning but cannot effectively work on the MIPL datasets, such as Newsgroups-{\scriptsize{MIPL}}, Birdsong-{\scriptsize{MIPL}}, and SIVAL-{\scriptsize{MIPL}}. This phenomenon indicates that it is necessary to propose tailored algorithms for solving the MIPL problems effectively. \\
\end{enumerate}

\subsection{Further Analyses}
\label{subsec:further_analyses}
\subsubsection{Exploration of Dirichlet Prior $\alpha_{\epsilon}$}
As defined in Eqs. (\ref{eq:init_weight},) (\ref{eq:update_weight}), and (\ref{eq:new_label}), the transformed labels are affected by the Dirichlet prior $\alpha_{\epsilon}$.
When $\alpha_{\epsilon}$ approaches $0$, the transformed labels and variances of non-candidate labels become negative infinity and positive infinity, respectively, which makes the Gaussian processes regression impossible. To avoid this issue, the Dirichlet prior plays a role in restricting the transformed labels of the non-candidate labels finite.
At the same time, the consequential labels and variances of candidate labels are negative values and positive ones that come near to $0$. During the iterations, the transformed results of the ground-truth labels are closer to $0$ than those of the false positive labels, and the differences between the ground-truth labels and the false positive ones become larger. 
\begin{figure}[!t]
\begin{overpic}[width=150mm]{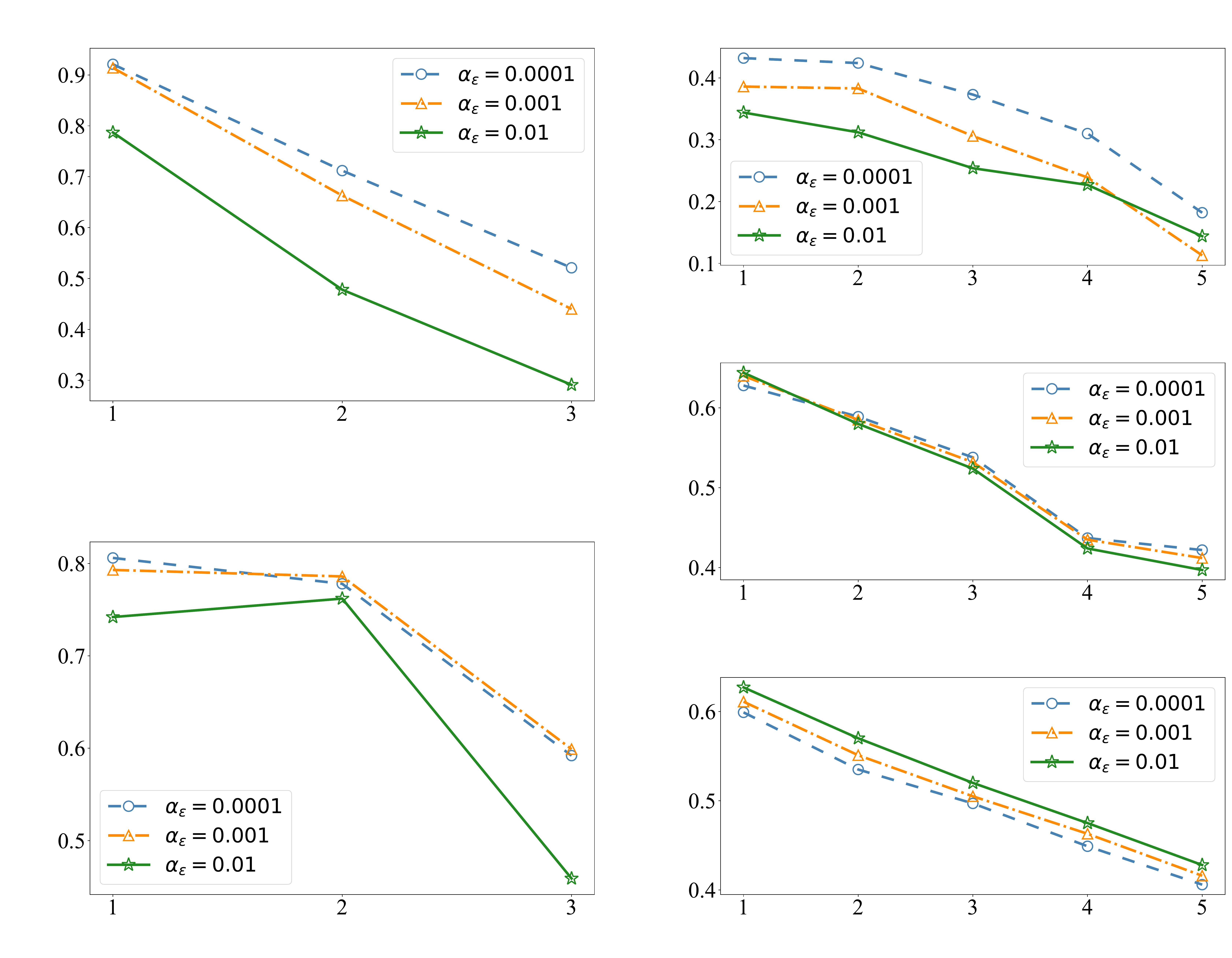}
\put(200, 745) {\footnotesize MNIST-{\tiny{MIPL}}}
\put(150, 415) {\footnotesize $r$ ($\#$ the false positive labels)}
\put(20, 555) {\footnotesize \rotatebox{90}{Accuracy}}

\put(195, 345) {\footnotesize FMNIST-{\tiny{MIPL}}}
\put(150, 10) {\footnotesize $r$ ($\#$ the false positive labels)}
\put(20, 170) {\footnotesize \rotatebox{90}{Accuracy}}

\put(690, 745) {\footnotesize Newsgroups-{\tiny{MIPL}}}
\put(645, 525) {\footnotesize $r$ ($\#$ the false positive labels)}
\put(535, 600) {\footnotesize \rotatebox{90}{Accuracy}}

\put(710, 490) {\footnotesize Birdsong-{\tiny{MIPL}}}
\put(645, 270) {\footnotesize $r$ ($\#$ the false positive labels)}
\put(535, 350) {\footnotesize \rotatebox{90}{Accuracy}}

\put(730, 235) {\footnotesize SIVAL-{\tiny{MIPL}}}
\put(645, 10) {\footnotesize $r$ ($\#$ the false positive labels)}
\put(535, 100) {\footnotesize \rotatebox{90}{Accuracy}}
\end{overpic}
\caption{Classification accuracy of {M\scriptsize{IPL}}G{\scriptsize{P}} on the MIPL datasets with varying $r$ and $\alpha_{\epsilon}$.}
\label{fig:epsilon}
\end{figure}
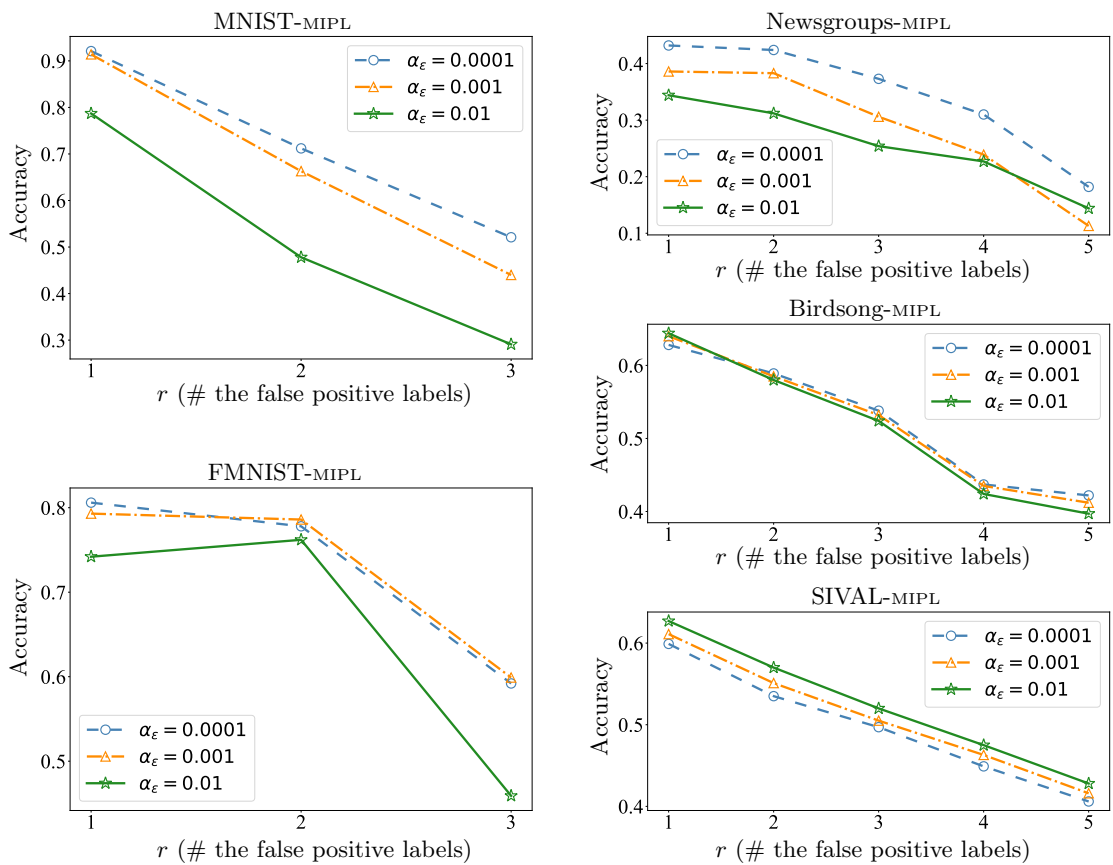

The classification accuracy of {M\scriptsize{IPL}}G{\scriptsize{P}} with the varying number false positive labels $r \in \{1,2,3,4,5\}$ and the different Dirichlet prior $\alpha_{\epsilon} \in \{0.0001,0.001,0.01\}$ on the MIPL datasets is shown in Figure \ref{fig:epsilon}. There are several observations: 
\begin{enumerate}
\setlength{\itemsep}{-10pt}
\setlength{\parsep}{0pt}
\setlength{\parskip}{0pt}
	\item[•] On MNIST-{\scriptsize{MIPL}} and Newsgroups-{\scriptsize{MIPL}} datasets, the smaller $\alpha_{\epsilon}$ can achieve better results, while the results are reversed on SIVAL-{\scriptsize{MIPL}} dataset. \\
	\item[•] On Birdsong-{\scriptsize{MIPL}}  datasets, the differences between the varying $\alpha_{\epsilon}$ are slight. Similarly, there is no obvious difference between the $\alpha_{\epsilon}=0.0001$ and $\alpha_{\epsilon}=0.001$ on FMNIST-{\scriptsize{MIPL}}  dataset. \\
	\item[•] Different datasets have diverse optimums of the Dirichlet prior, which are determined by the characteristics of the datasets themselves.
\end{enumerate}
In our experiments on {M\scriptsize{IPL}}G{\scriptsize{P}} and the two variants, we set $\alpha_{\epsilon}$ to $0.0001$ which can perform satisfactorily on all datasets.

\section{Conclusion}
\label{sec:conclusion}
In this paper, we formalize a novel learning framework named multi-instance partial-label learning (MIPL), where each training sample is associated with not only multiple instances but also a candidate label set that contains one ground-truth label and some false positive labels. Although the MIPL problems widely exist in many real-world applications, to the best of our knowledge, {M\scriptsize{IPL}}G{\scriptsize{P}} proposed in this paper is the first tailored MIPL algorithm. Specifically, {M\scriptsize{IPL}}G{\scriptsize{P}} transforms the candidate label sets from the augmented label space into a logarithmic space, yielding a Gaussian likelihood and transforming the classification problem into a regression problem. To solve the regression problem, {M\scriptsize{IPL}}G{\scriptsize{P}} induces an efficient Gaussian processes model with GPU accelerations. Extensive comparative studies validate that existing multi-instance and partial-label algorithms are not able to handle the MIPL problems, and {M\scriptsize{IPL}}G{\scriptsize{P}} performs significantly better than other algorithms under the MIPL setting. In the future, there are many directions to explore. For example, exploiting the instance-candidate label dependencies or exploring the theoretical properties of MIPL.


\bibliographystyle{named}
\bibliography{miplgp}

\newpage
\section*{Appendix A. The MIPL Datasets}
\label{sec:Appendix A}
In this section, we provide the details of the MIPL datasets, i.e., MNIST-{\scriptsize{MIPL}}, FMNIST-{\scriptsize{MIPL}}, Newsgroups-{\scriptsize{MIPL}}, Birdsong-{\scriptsize{MIPL}}, and SIVAL-{\scriptsize{MIPL}}.

For MNIST-{\scriptsize{MIPL}}, FMNIST-{\scriptsize{MIPL}}, and Newsgroups-{\scriptsize{MIPL}} datasets, we need to choose the targeted class labels and the reserved class labels to provide each multi-instance bag with positive instances and negative ones. For MNIST-{\scriptsize{MIPL}} dataset, we extract $\{0,2,4,6,8\}$ as five target classes for providing the positive instances according to the corresponding class and draw all negative ones from the reserved classes $\{1,3,5,7,9\}$ randomly. For FMNIST-{\scriptsize{MIPL}} dataset, the targeted class labels and the reserved class labels are $\{$\emph{T-shirt, Trouser, Coat, Sneaker, Bag}$\}$ and $\{$\emph{Pullover, Dress, Sandal, Shirt, Ankle boot}$\}$, respectively. 
Newsgroups-{\scriptsize{MIPL}}, the dataset is widely used in binary multi-instance learning, where each instance is represented by the top $200$ TF-IDF features, and each positive bag contains $3\%$ positive instances drawn from the target class. Similarly, we represent each instance by the top $200$ TF-IDF features in Newsgroups-{\scriptsize{MIPL}}, and Table \ref{tab:20news} summarizes the targeted class labels and the reserved class labels of Newsgroups-{\scriptsize{MIPL}} dataset.
\begin{table}[!b]
  \caption{The targeted class labels and reserved class labels of Newsgroups-{\scriptsize{MIPL}} dataset.}
  \label{tab:20news}
  \centering
    \begin{tabular}{ll}
    \hline   \hline
    Targeted class labels & Reserved class labels \\
    \hline
    alt.atheism & comp.graphics \\
    comp.os.ms-windows.misc & comp.sys.ibm.pc.hardware \\
    comp.sys.mac.hardware & comp.windows.x \\
    misc.forsale & rec.motorcycles \\
    rec.autos & rec.sport.baseball \\
    rec.sport.hockey & sci.crypt \\
    sci.med & sci.electronics \\
    sci.space & talk.politics.guns \\
    soc.religion.christian & talk.politics.misc \\
    talk.politics.mideast & talk.religion.misc \\
    \hline \hline
    \end{tabular}
\end{table}

The Birdsong dataset is proposed in multi-instance multi-label learning, which contains $548$ multi-instance bags totalling $10232$ instances. Each instance is represented by a $38$-dimensional feature vector and associated with a single label, which is chosen from $13$ targeted class labels or $1$ negative class label.
In Birdsong-{\scriptsize{MIPL}}, the $13$ targeted class labels and the negative class label are regarded as the targeted class labels and the reserved class label, respectively.

SIVAL is a multi-instance learning dataset for content-based image retrieval with $1500$ images. Each image is a multi-instance bag, which is associated with one of $25$ class labels and consisted of $31$ or $32$ instances. In addition, each instance is represented by a $30$-dimensional feature vector. To yield the SIVAL-{\scriptsize{MIPL}}, we only need to generate the false positive labels for each image. Specifically, we treat the $25$ class labels as the targeted class labels and sample $r$ false positive labels from the targeted class labels excluding the ground-truth label randomly.

\end{document}